**Title:** Flexoskeleton printing for versatile insect-inspired robots
**Short title:** Flexoskeleton printing for versatile insect-inspired robots

**Authors:** M. Jiang, Z. Zhou, N. Gravish*

**Affiliations:**
[1]Department of Mechanical and Aerospace Engineering
University of California, San Diego.
*Correspondence to: ngravish@eng.ucsd.edu

**One Sentence Summary:** 3D printing onto heated thermoplastic layers enable rapid fabrication of flexible and resilient robots inspired by the multi-stiffness properties of the insect exoskeleton.

**Abstract:** One of the many secrets to the success and prevalence of insects is their versatile, robust, and complex exoskeleton morphology. A fundamental challenge in insect-inspired robotics has been the fabrication of robotic exoskeletons that can match the complexity of exoskeleton structural mechanics. Hybrid robots composed of rigid and soft elements have previously required access to expensive multi-material 3D printers, multi-step casting and machining processes, or limited material choice when using consumer grade fabrication methods. Here we introduce a new design and fabrication process to rapidly construct flexible exoskeleton-inspired robots called flexoskeleton printing. We modify a consumer grade fused deposition material (FDM) 3D printer to deposit filament directly onto a heated thermoplastic base layer which provides extremely strong bond strength between the deposited material and the inextensible, flexible base layer. This process significantly improves the fatigue resistance of printed components and enables a new class of insect-inspired robot morphologies. We demonstrate these capabilities through design and testing of a wide library of canonical flexoskeleton elements; ultimately leading to the integration of elements into a flexoskeleton walking legged robot.

**MAIN TEXT**

**Introduction**

The huge diversity of body morphologies and locomotion capabilities in the insect world have long served as inspiration for the design and control of flying (*1*, *2*), swimming (*3*, *4*), and walking robots (*5*–*7*). A defining feature of insects (and more broadly all arthropods) is their external skeleton, called an exoskeleton, which must serve multiple roles including structural support, joint flexibility, joint and body protection, and providing functional surface features for sensing, grasping, and adhesion(*8*) (Fig. 1). The exoskeleton of all insects is a continuous sheath encompassing the animal, largely formed from two materials: chitin networks that are embedded within cuticular proteins (*9*, *10*). Variation in exoskeleton stiffness (and other mechanical properties) occurs within the continuum of the exoskeleton to distinguish joints, struts, and continuously flexible regions (Fig. 1a). Both stiffness gradients, and discrete changes in stiffness, are controlled by variations in exoskeleton thickness, scleritization, and geometry. Critically, the mobility and functional capabilities of insect limbs are determined by this arrangement of rigid, soft, and graded stiffness elements. The insect exoskeleton truly embodies a hybrid structure of rigid and soft mechanical elements (*9*, *11*, *12*).

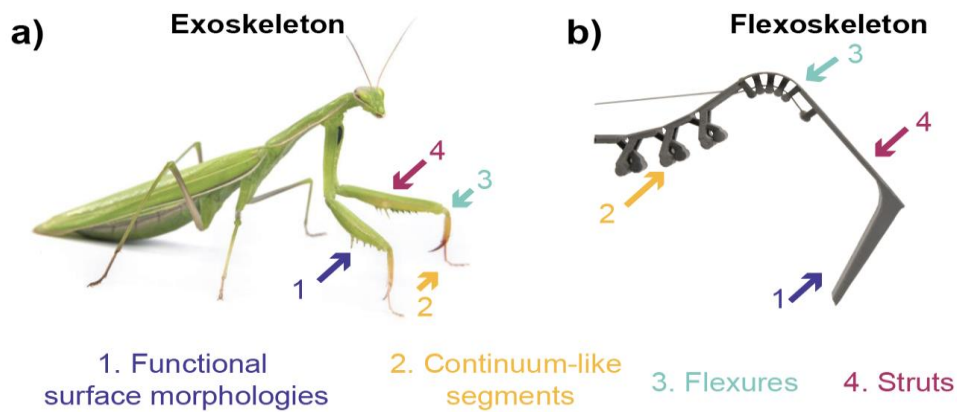

Figure 1. a) The exoskeleton of an insect provides protection and serves to support locomotion through structural and flexural regions. b) In this paper we introduce a new method for fabricating exoskeleton-inspired robots that seek to embody the four principles of the exoskeleton.

Animals, including insects, have long served as inspiration for robotics(*3*, *13–15*). However, until recently, bio-inspired robots tended to look like their rigid industrial robot counterparts, with legs and joints built from rigid links and stiff, high-gear ratio motors (*16*, *17*). More recently roboticists have begun adopting a bio-inspired approach to robot structure, intentionally including body and limb compliance in robot designs (*15*, *18–21*). New fabrication methods were developed to support this new direction in bio-inspired robotics including shape-deposition manufacturing (*22–26*), multi-material 3D printing (*27*, *28*), laser-cutting and lamination (*1*, *29–31*), and mold-casting (*32*, *33*). However, the use of these techniques has often relied on access to expensive and time consuming fabrication tools with multi-step processes and limited materials selection.

In this paper we introduce a novel fabrication process called "flexoskeleton" printing to 3D print flexible and resilient robot exoskeletons for insect-inspired robots. This method uses low-cost fused deposition modeling (FDM) 3D printers and standard rigid filament materials (ABS/PLA) that is readily accessible. The fundamental advance of our method relies on 3D printing rigid filaments directly onto a heated thermoplastic film which provides a flexible, yet strong base layer to the deposited material. This fabrication method enables precise control of the arrangement and stiffness properties of joints and struts within the continuum structure of the robot body and its ease of use will enable wide adoption and significantly reduced fabrication times for bio-inspired robotics.

Below we describe in detail the flexoskeleton fabrication process. We begin with a description of the fabrication process and experiments to demonstrate the robustness and fatigue resistance of these structures. We then present a library of flexoskeleton components that enable control of joint stiffness and bending limits. Integration of multiple joint elements into a single structure enables complex motion from multi-jointed legs that can be optimized for robot walking behavior. Lastly, we demonstrate the capabilities of this rapid design and fabrication process by building and testing a quadruped flexoskeleton walking robot.

**Results:**

*Flexoskeleton printing*

The flexoskeleton printing process involves a small modification to the standard fused-deposition modeling (FDM) approach to 3D printing. In a standard consumer grade FDM printer a plastic filament such as ABS or PLA is extruded through the aperture of a heated nozzle and deposited onto a flat print surface. Many consumer FDM printers enable control of the print surface temperature as well, so that the printed material can resist warping from thermal gradients during the print process. In the flexoskeleton printing process we adhere a thin sheet of polycarbonate, a thermoplastic that can be softened and molded under heating, to the heated bed upon which we directly print. By heating the polycarbonate (further referred to as PC) we are able to achieve very strong adhesion between the 3D printed material and this base layer which enables the printing of resilient flexible structures on standard consumer FDM printers.

The print process begins by securing the base layer PC film on to the heated bed surface using a standard adhesive such as a washable glue stick (See Fig.2 and SI Movie 1). To further reduce warping of the PC we additionally tape down the edges using high temperature masking tape. We next allow the bed temperature to reach the desired temperature, typically between 80 - 100°C and once the bed temperature is stable the print process begins. A common variable to control in FDM printing is the Z-offset between the print nozzle and the bed height for the first layer of printing. To allow the first deposited layer to achieve close contact with the PC layer, and to create enough contact pressure for good bonding we set a relatively small Z-offset, between 0.01 - 0.03mm. After the full print operation is finished we first allow the heated bed and part to cool which depending on the size of the part can be 5-20 minutes. Once the bed has cooled we peel off the PC layer, including the bonded 3D printed components, and we remove the excess PC layer as the design dictates (Figure 2a,b and c). We currently manually trim the PC layer with a cutting tool such as scissors or a razor, however future flexoskeleton processes may integrate automated pre-cutting of the PC film using a vinyl cutter or laser cutter (as applicable to the base layer material).

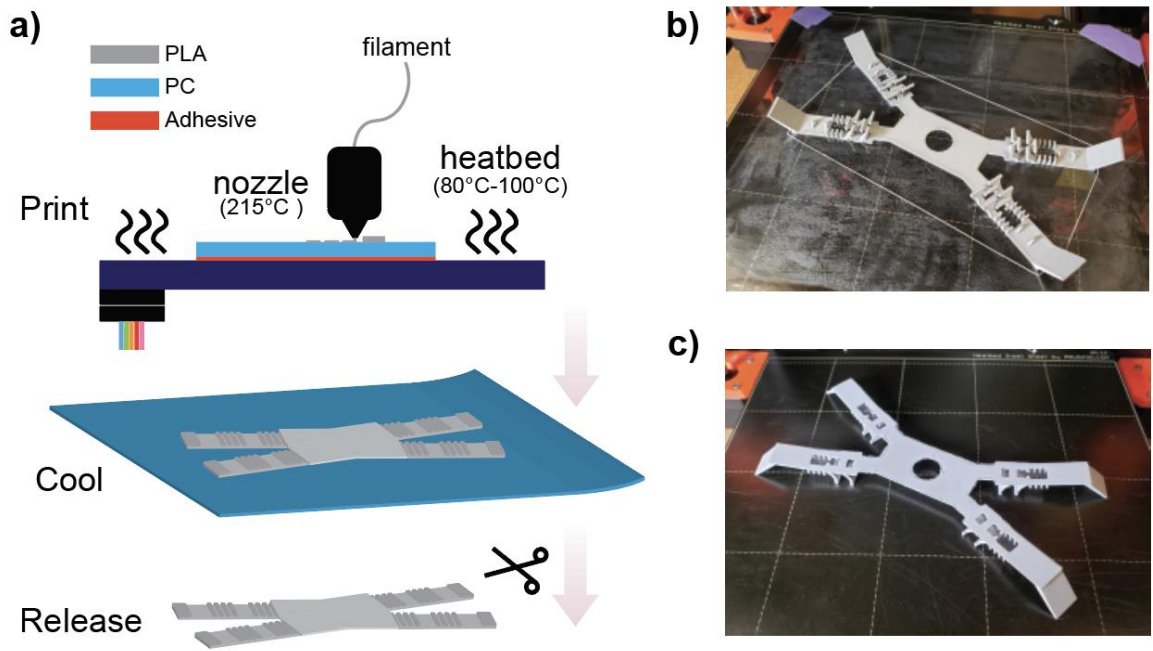

Figure 2. Fabrication method. a) Schematic of the flexoskeleton printing process. Printer filament (PLA or ABS) is deposited on a heated layer of thermoplastic (PC). The PC is adhered to the surface of the print bed to remain flat during printing. Once printing is completed the base layer and printed material are allowed to cool. After cooling the part is released from the PC sheet by cutting. b) An example of a four-legged robot immediately after printing on clear PC layer. c) A four legged robot after release from the PC layer.

*Delamination and fatigue resistance*

High bonding strength between the constitutive materials in either laminate fabrication (*34*) or multi-material 3D printing (*35*) is one of the most desirable mechanical properties to improve component lifespan and usability. For multi-layered laminate robots, sheet adhesives (double-sided tape, thermoset adhesives) and liquid adhesive (epoxies, cyanoacrylate) have been extensively used for bonding (*36*). While these adhesives are often extremely strong, application requires a multi-step alignment and bonding process. Multi-material 3D printing also relies on the bonding strength between dissimilar materials that are printed into the continuum structure. High-end multi-material printers are often able to achieve strong bonding performance between rigid and soft materials, however this comes at the expense of long print times, expensive print materials, and expensive printers. Alternatively, consumer grade multi-material printing capabilities are emerging but suffer similar challenges in print time with relatively poor bond strength between dissimilar materials (*37*, *38*).

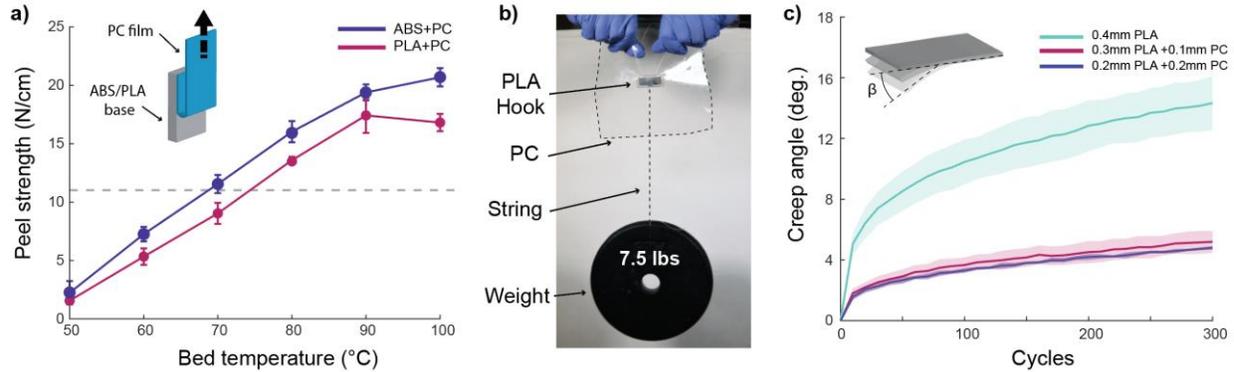

Figure 3. Examples of resistance to delamination and fatigue. a) The bed temperature determines the delamination peel strength between the PC layer and the 3D printed features. Peel strength in 180 degree peel tests increased to a maximum at 90-100°C. The dashed line indicates the peel strength (90 degree peel test) between a standard industrial acrylic adhesive on ABS for reference. b) Proper print settings enable strong resistance to delamination between 3D printed features and the PC layer. Here a 12x34 mm rectangle printed onto the PC layer is able to hold up a 7.5 lb weight. c) Cyclic bend tests of constant thickness flexures printed with (red and blue) and without (green) show significantly less creep over time.

The bonding process for flexoskeleton printing does not require additional adhesives or curing agents as the filament will be directly bonded to the base PC layer during extrusion. The bond between the first deposited layer and the base PC layer can be regarded as a thermal bonding process between two thermoplastics in which high bonding pressure and temperature are crucial for an overall high bonding strength (*39*). The quality of the bond is thus likely to be sensitive to the bed temperature, while the nozzle temperature should be kept fixed to maintain print quality (such as print resolution and stringing). To determine the optimal thermoplastic heating parameters we printed peel test samples with uniform geometry onto PC films (0.1mm thickness) under varying heatbed temperatures. We conducted 180° peel tests by peeling the PC layer from the printed sample using an Instron 3367. We measured the peel force during delamination and report the peel strength as a function of bed temperature (See Materials and Methods for further information). We find that for both ABS and PLA print material, the adhesion strength to PC is strongly dependent on heatbed temperature. We observed that the peak peel strength for both PLA and ABS occurred for heatbed temperatures between 90 - 100°C (Figure 3a). To give a tangible understanding of the high peel strength, we conducted a simple demonstration where a rectangular printed hook (surface area 12x34mm) was able to pick up a 7.5 lb weight using a string without being delaminated (Figure 3b). Furthermore, we also find that the peel strength can exceed that of commercially available high-bond acrylic sheet adhesives (3M high bond transfer tape, 90° peel test, 11.2N/cm onto ABS) demonstrating the strong delamination resistance of flexoskeleton components.

One critical challenge facing FDM 3D printed components from standard filaments (ABS/PLA) is its low fatigue resistance (*40*, *41*). Consumer grade printed components typically will quickly yield or break under cyclic loading conditions, and thus are not recommended for using as long-term bending flexures especially with large bending range (*42*, *43*). Flexoskeleton printed components on the other hand may overcome rapid fatigue and failure as the PC base layer acts as a tension-resistive protective layer. Compared with ABS and PLA filament, PC film has high flexural resilience, good impact resistance and toughness, and high tensile strength. Thus, the addition of a PC layer can reduce the amount of plastic deformation and fatigue that shallow layers of FDM printed components typically experience. To test this hypothesis, we fabricated flexible beams with uniform rectangular geometries [32(L)x22(W)x0.4(T)mm] under three conditions; a standard printed control sample with no PC layer, and two flexoskeleton beams with PC layers of 0.1 and 0.2mm. We mounted the samples on a cyclic loading apparatus that bent the beams unidirectionally between a rest position and desired bend condition. In this test we bent each beam to a constant stress state and maintained this position for 10 seconds to simulate scenarios where robot legs will be bent and held in place for load support. We measured the creep angle of the beam by taking an image of the unloaded beam deflection angle as measured from the neutral position before testing. We find that by adding a PC layer we are able to reduce the creep deformation of 3D printed beams by 70% over a 300 load cycle period (Fig. 3c). The thickness of the PC layer for the two samples did not further contribute to the creep behavior of these beams during cyclic bending moments. This experiment demonstrates how the flexoskeleton printing process can enable direct printing of flexures and structures into a single continuum for bio-inspired robotic exoskeletons.

*A library of programmable stiffness sub-components*

Having demonstrated the viability of flexoskeleton printing for creating resilient flexure elements we now explore how the morphological features of the printed layer can be modulated to control bending properties. First, we investigate how flexure stiffness can be controlled by printing simple linear patterns. One way to modulate the bending stiffness of the flexure is by increasing the printing thickness of the deposited materials. However, as most consumer grade printers can only print layers at a poor resolution (approximately 0.1 - 0.3mm layer resolution), depositing uniform layers may not enable fine scale control of bending stiffness. Instead, a simple design principle for stiffness control is demonstrated in which linear patterns of high and low segments are printed across the flexure region (Figure 4a). Here, the thickness of the segment is defined as 'feature height' and the ratio of the raised feature's width versus period is defined as the 'feature width ratio'. We then controlled these 2 design parameters separately and performed linear stiffness tests by using a custom built rotary testing stage and measuring stiffness under small bending angle (See Materials and Methods for further information). As shown in Figure 4a, changing the feature height results in relatively poor control of joint stiffness since the stiffness curve eventually reaches a plateau as height increases. However, changing the feature width ratio provides an effective method for control of flexure bending stiffness (Figure 4b). The predicted stiffness for all samples (dashed line) are based on a simple Euler Bernoulli beam theory model that only uses geometric and material parameters (*48*).

In addition to control of joint stiffness, many animals and robots possess joint stops to limit a joint's range of motion. Here we demonstrate two types of flexoskeleton joint limits which are both based on the principle of jamming between extruded features at a desired bend angle. Flexional joint limits are composed of vertical pillars with a circular end that serve to jam together and significantly stiffen the joint (Figure 4c). The distance between the center of the head to the base layer is defined as the 'feature height' and a simple geometrical model (See SI) dependent on these parameters is provided. Here we report programmable joint limits by varying the feature height (Figure 4c right). By reversing the order of the adjacent jammable features for every two stand outs, we reversed the direction in which the jamming happens and thus created extensional joint limits stiffening the joint at large extensions of the joint. Here we defined the feature length as the diagonal length of the geometrical stand outs and measured jamming angles versus this control parameter. In all instances we find good agreement between measurements of the jamming angle and a simple geometrical model of flexure bending (see SI for further information).

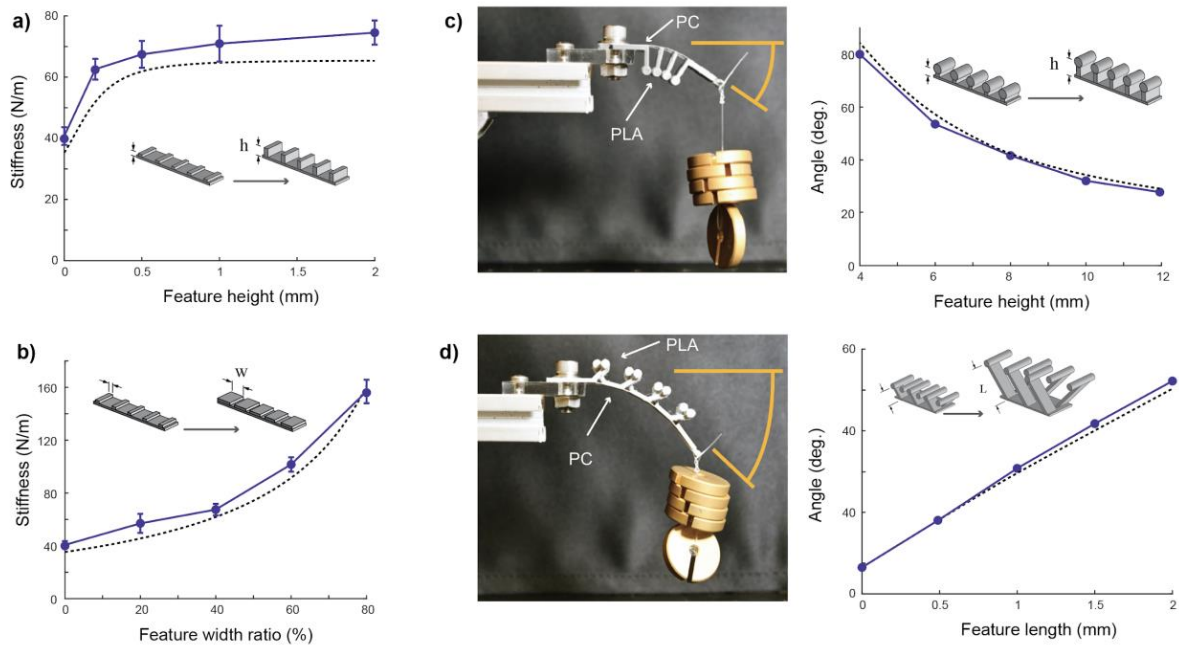

Figure 4. Flexoskeleton design features. a-b) Control of flexure mechanical stiffness is controlled by the width and height of rib features printed on the flexure. c-d) To limit the angular range of printed flexure joints we design jamming features that stop joint motion at a desired angle. Joint limits can be introduced in both the flexion (c) and extension (d) directions. Dashed lines represent theoretical model predictions (see SI sections 1 and 2).

*Complex limb motions determined by leg mechanical properties*

As explored in the previous section, adding geometric features to the printed layers can enable a wide variety of flexure bending mechanical properties, such as variable bending stiffness, and joint limits. These features can thus be integrated into a robotic limb for programmable limb motions. As a demonstration of generating complex motions in underactuated flexoskeleton limbs, we designed a flexoskeleton leg composed of two flexural joints: a flexion joint for limb contraction and an extension joint for limb lifting the foot tip. For each joint, we prescribed the passive joint stiffness (determined by the segmented patterns), and joint limits (flexion and extension) and used a single tendon to actuate the two joints by routing it above the extension joint and below the flexion joint (Figure 5a). We present the design of the hind limb as an example, to achieve a non-repeatable limb cycle we printed higher passive joint stiffness for the flexion joint but lower passive joint stiffness for the extension joint. This enables the leg to lift first then contract back and downwards for a foot displacement without touching the ground (Figure 5b). As the tendon releases, asymmetric friction at the toe and tendon cause the foot to generate a push motion against the ground and thus enabling hysteretic foot motion via a single tendon. The quality of the push stroke is controlled by the jamming hinge angles and stiffnesses. As an exploration of this we changed the design of the extensional joint limits, observing different foot trajectories (Figure 5c) and measuring the limb curvature and shape change (Figure 5d) and stroke properties (Figure 5e). Tracking the continuous curvature of the limb during the actuation cycle highlights how flexoskeleton limbs behave as a continuum structure with a gradient in shape and curvature (Figure 5d). This is in stark contrast to more traditional link-joint limb designs in which the curvature would be observed as a delta function at each joint.

The stroke distance as well as the stroke ratio (defined as stroke distance per unit tendon pull distance) is controlled by the properties of the limb hinges, and in Figure 5e we highlight how changing the jamming angle of the extensional hinge alone can enhance or degrade the limb stroke quality. We find that by having different designs of the jamming morphology one can achieve different stroke distances with no big change of the stroke ratio, whereas changing the programmable passive joint stiffness allows for different walk sequences and thus the functionality of the limb. The design of the front limb can be achieved by having the opposite stiffness distribution as in the hind limb which enables a pull-lift cycle of movement.

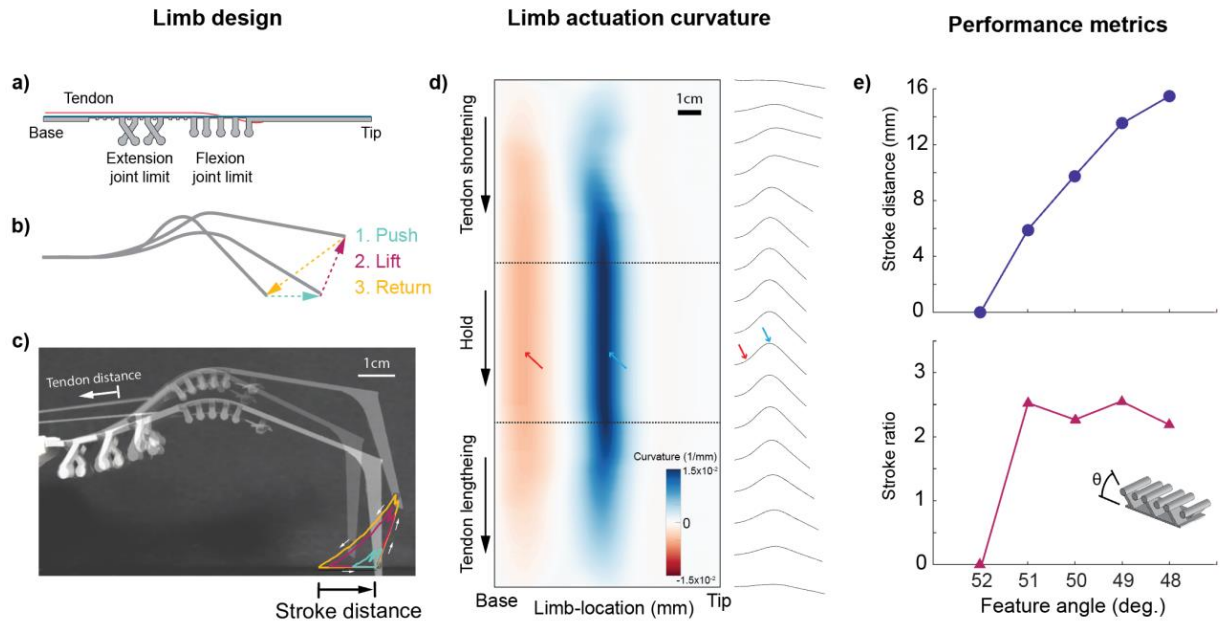

Figure 5. Limb motion control through mechanically programmable joint properties. a) Design of one robotic limb, composed of both extension joint and flexion joint with programmable linear stiffness and joint limits. b) The motion sequence of a hind leg from pulling and releasing of one tendon. c) Foot trajectories tracked from within one limb motion cycle with different jamming feature angles (yellow: 48 deg., red: 49 deg. and green: 51 deg.). d) Heat map of the leg actuation curvature (right) and the shape of the continuum limb outlier (left) changing as a function of the tendon state. e) Stroke distance (top) and stroke ratio (bottom), defined as stroke distance per unit tendon pull distance as measured from different feature angles (from 48 deg. to 52 deg.).

*Walking performance of a flexoskeleton walking robot*

As a demonstration of the walking capabilities of a rapidly prototyped flexoskeleton robot, we built a tendon driven four-legged flexoskeleton walking robot. The robot is designed and assembled using all flexoskeleton printed limbs and chassis, and is actuated by 4 micro servos (Figure 6a). We chose the front limbs and hind limbs to have similar bending stiffness properties, however the high and low stiffness of the flexional and extensional joints were reversed between rear and front limbs to reverse the stroke direction. Each limb is then inserted into the robot chassis (body) and connected with one micro servo (Tower Pro SG92R) through a capstan and tendon. Having a 'plug and play' limb design enables us to rapidly swap any robot limb to suit for different walking terrains (such as smooth or rough substrates). Note that the robot can also be printed within one print if certain joint parameters are pre-programmed to meet with specific walking requirements. For robot walking, we first started the robot by applying pre-tension in the tendon to support the whole body weight (78g) while standing. The joint limits can further help the robot support the stationary body weight by stiffening the joint at the extreme flexion or extension angles. We then programmed each bi-pod walking gait with 2 diagonal pairs (e.g. front left and hind right as one pair) walking out of phase with the same frequency (See supplementary movie S2). The speed of the robot was measured against different driving frequencies as we tracked the robot walking on a flat and smooth surface (paper substrate).

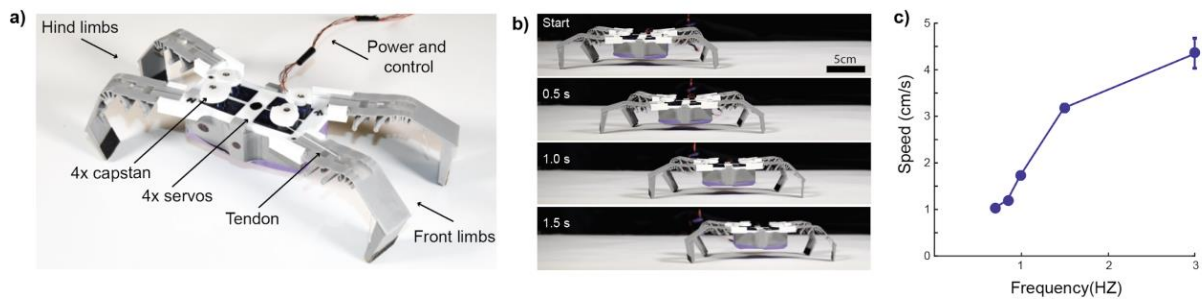

Figure 6. A walking quadruped flexoskseleton robot. a) Each limb is actuated by a single servo and tendon. Off-board power and control is provided through a tether. b) A side view of a walking sequence. c) Speed-frequency relationship of the walking performance of the robot.

**Conclusion:**

The arthropod exoskeleton serves both as a protective exterior, and as a mechanical transmission that routes power from muscles to limbs. The exoskeleton is a multi-material continuum in which rigid and soft tissues are organized in complex three-dimensional arrangements. Critically, the arrangement of these rigid and soft regions form functional mechanical systems such as linkages (*44*), springs (*45*, *46*), and even gears (*47*). Robot morphologies inspired by the insect exoskeleton may enable new multi-functional robots and further help us understand how complex arrangements of compliant elements can enable power transmission and control of biological locomotion.

Inspired by the insect exoskeleton, in the work described here we present a new fabrication process called "flexoskeleton" printing that enables rapid and accessible fabrication of hybrid rigid/soft robots. Critically, this fabrication approach is extremely accessible to both novice and expert users and does not require exorbitant material or equipment costs. The approach we have developed relies heavily on the interrelationships between three dimensional geometry of surface features and their contributions to the local mechanical properties of that component. We envision this method will enable a new class of bio-inspired robots with focus on the interrelationships between mechanical design and locomotion.

**List of Supplementary Materials:**

Materials and Methods

Figs. S1 to S2

**Citations:**

**Acknowledgments:** We thank the members of the Gravish Lab including Wei Zhou, Shivam Chopra, and Glenna Clifton for helpful project comments. **Funding:** Funding support was provided through the Mechanical & Aerospace Engineering Department at UCSD. **Author contributions:** Authors NG and MJ contributed to the conceptualization, design, and experiments in this work. Author MJ and ZZ contributed to mechanical characterization and robot design and fabrication work. All authors contributed to writing. **Competing interests:** All authors declare no competing interests. **Data and materials availability:** Data will be made available on Data Dryad

**Supplementary Materials for**

**Title:** Flexoskeleton printing for versatile insect-inspired robots
**Authors:** M. Jiang, Z. Zhou, N. Gravish*
Correspondence to: ngravish@eng.ucsd.edu

**This PDF file includes:**

Materials and Methods

Figs. S1 to S2

**SUPPLEMENTARY MATERIAL**

**Materials and Methods**

*Flexoskeleton 3D printing process*

The fabrication of flexoskeleton components relies on a simple modification of the standard fused-deposition-modeling (FDM) 3D printing process. The fundamental requirement of this process is an FDM printer with a heated bed and temperature control. We have performed all our printing using either the Prusa i3 MK3S and the LulzBot Taz 6. We have demonstrated successful flexoskeleton printing with both ABS and PLA filament (HATCHBOX, diameter 1.75mm) printed directly onto a thermoplastic backing film (PC, Polycarbonate, 0.1mm - 0.2mm). The fabrication process is as follows: first, the heatbed is preheated to 80 - 100°C depending on the bonding requirement (See Fig. 2a and SI Movie 1). We then secured the backing film on the heat surface using washable glue (gluestick) and taped down the edges of the film using high temperature masking tapes. The print was started with a preset Z-offset (0.01 - 0.03mm above the film surface) to allow the 1st deposited layer in close contact with the base film. The nozzle was heated up to 215°C for PLA and 240°C for ABS. After the prints finished, we peeled off the entire sheet and hand cut extra base film using scissors as shown in Figure 2a.

*Mechanical characterization*

The peel strength between the PC and the deposited material was measured based on the 180 degree peel test using Instron 3367. For preparing each test sample, we printed a block of size 40(L)x12(W)x2(T)mm PLA/ABS (215°C/240°C nozzle temperature) on top of a sheet of 0.1mm thickness PC film under different heatbed temperatures (50~100°C). As the prints finished, we peeled off the whole sheet and cut all samples to size and pre-peeled the PC off from one end. The free end of PC was then clamped by the vise as well as one side of PLA/ABS edge. The machine then pulled the PC film against the PLA/ABS base with a 180 degree peel angle and the results are further calculated as the average peel force (over a 5mm peel distance) divided by the peel width (N/cm).

The fatigue test was conducted by applying a cyclic bending load with constant maximum stress onto flexure samples. Here we maintained the same sample thickness while changing the relative thickness of the printed material and the PC layer. The cyclic load is created by using a 3D printed L-shape rod (PLA, 100% infill, 0.1mm layer resolution) driven by a stepper motor (VEXTA, PK546PMB) which rotated the free end of the test samples (32(L)x22(W)x0.4mm(T)) with the base mounted on a fixture (coaxial with the motor axis). The load cycle was composed of a unidirectional bend (30 degrees/second) and a 10-second pause (holding in place) at the maximum bending amplitude to simulate the applicational scenarios where the joint will be bent and held for load support within a duration of time. To keep the maximum bending stress the same, we adjusted the maximum bending amplitudes among all samples to keep the maximum bending torque constant. We then used a webcam placed above the test samples to record the creep angles between each load cycle.

The measurement of the linear flexure (flexoskeleton) stiffness was based on a custom motorized stage and force sensor setup. We mounted a load cell (Futek, LSB 200) on a rotary disk stage which applied normal force directly onto the free end of the flexure. The free end was loaded along a circular path while force was measured. The test samples are 30(L)x44(W)mm with a base thickness of 0.3mm (0.1mm PC film+0.2mm base PLA). Two parameters of the test samples were varied: feature height ranging from (0.2~2mm) and feature width ranging from 0 to 4mm (0% to 80%) within a 5mm feature cycle (one repeatable line segment). The samples were mounted with their base fixed at the center axis of the rotary stage. The test is then begun by driving the stage at 0.5 degrees/second in a 4 degree bending range with the segmented pattern on the flexure facing towards the load cell. The linear stiffness is then calculated as the force applied divided by the circular path travelled (N/m).

*Robot details*

Our robot is designed as a four-legged, tendon-driven walking robot, with flexoskeleton printing for all leg and chassis components enabling a low cost, rapidly prototyped, versatile walking robot. We used 4 micro servo motors (TowerPro, SG92R), powered by an arduino to control each leg individually. Each leg is driven by a micro servo motor using one tendon (fishing wire), which is pre-tensioned to support the whole body weight when the legs are stationary. To enable leg replacement the robot was printed in two parts: individual legs and a single chassis. The legs consist of a 70mm length flexoskeleton limb with two joints: one flexion, and one extension. We designed flexional limit that can be jammed at 90°, with a total joint length of 10mm and extensional limit jamming at 20° with a total joint length of 22mm. The function of the hind legs and front legs is different because of the requirements of the directional power stroke. For the front legs, the passive linear stiffness is determined by both the base PC film (0.2mm) as well as a printed base PLA layer. For the hind leg design, we made the extension joint (0.2mm PC+ 0.3mm PLA) a bit stiffer than the flexional joint (0.2mmPC + 0.1mm PLA) where as in the front leg design, the case was reversed. Such a design can help the hind leg to generate a lift-push gait cycle with the front leg doing a pull-lift gait cycle. Each leg is printed within 30 minutes with the total print time for the whole robot around 3 hours using one 3D printer. The layer height is set as 0.2mm with an infill of 30% for the robot chassis.

For the robot walking cycle, we implemented a 'trot-like' gait where diagonal pairs of the 4 legs (e.g. hind right and front left) were driven in phase with 2 diagonal pairs driven out of phase (alternating during one walk cycle). We then measured the walking speed under different motor driving frequencies on a flat surface with paper substrate. The speed is recorded as an average speed over a 0.8 meter walking distance using a camera for observation.

**Model Derivation**

*Joint limit prediction model:*

Flexional joint:

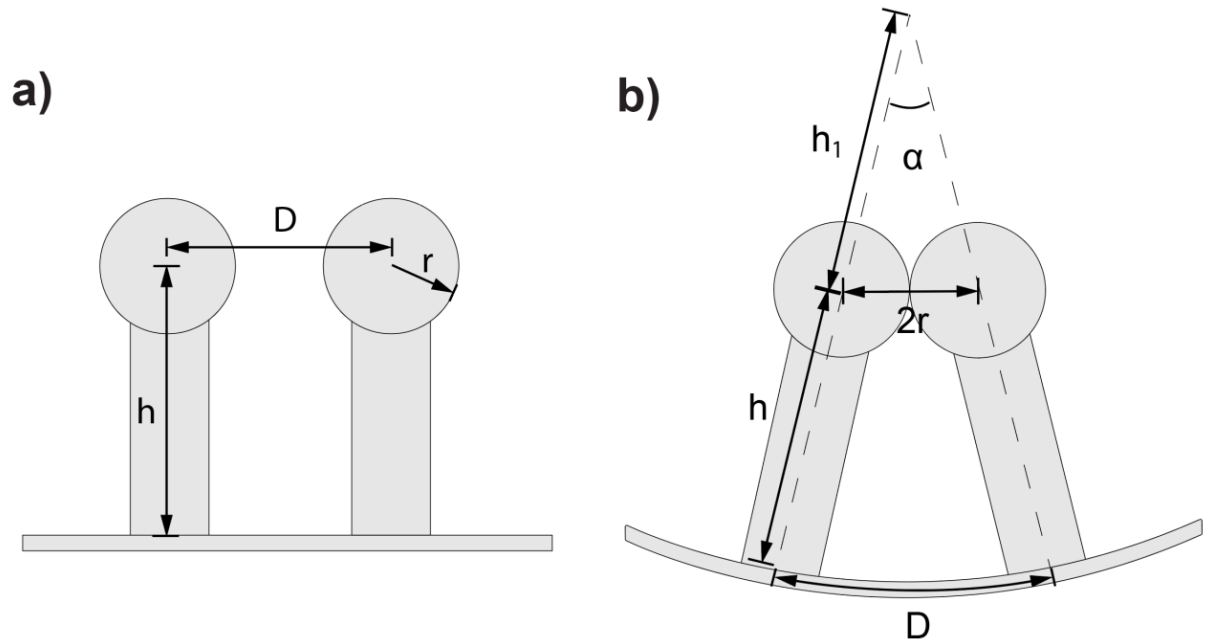

Figure S1. illustration of flexional joint limit. a) before jamming. b) after jamming.

Here the joint limit is defined as the jamming angle α (Figure S1b), that is formed by the mushroom-like jamming features when the flexure is bent at the extreme angle. First, since the base film is inextensible, we can assume that the arc length of the bent flexure between 2 adjacent jamming features is D (Figure S1b).

The jamming angle α can be solved as follows:

$$\alpha = \frac{D}{(h_1 + h)} \qquad (1)$$

$$\sin\left(\frac{\alpha}{2}\right) = \frac{r}{h_1} \qquad (2)$$

Since r, h and D are all know parameters, we can then solve α based on the above 2 equations. In the paper, the samples we used are h=4,6,8,10,12mm; r=2mm; D=6mm..

Extensional joint:

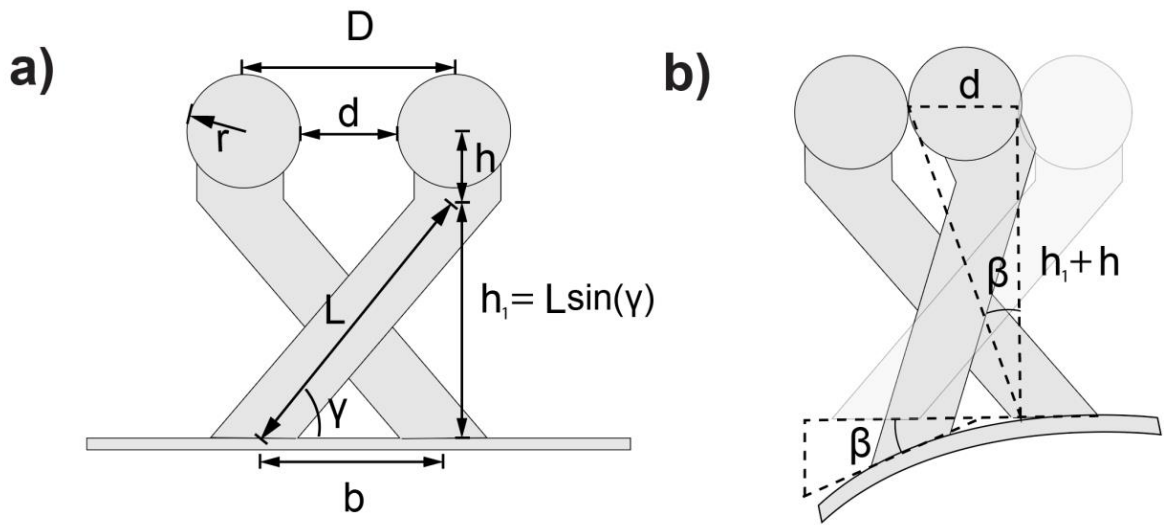

Figure S2. illustration of extensional joint limit. a) before jamming. b) after jamming.

The calculation of the jamming angle of the extensional joint limit can be considered as one jamming feature being rotated against the other jamming feature with the origin fixed at the root of the jamming feature, shown in Figure S2b.

The Jamming angle $\beta$ can be solved as:

$$\beta = \frac{d}{(h_1+h)} \qquad (3)$$

where d is determined by

$$d = D - 2r \qquad (4)$$

D is determined by

$$D = 2L\cos(\gamma) - b \quad (5)$$

And h1 is determined by

$$h_1 = L\sin(\gamma) \quad (6)$$

In our angular measurement of the joint limits, the samples we printed are

L=6.50,6.75,7.00,7.25,7.50mm; b=5.4mm; r=1.8mm; h=2mm; $\gamma$=45deg.